\documentclass{article}

\usepackage[preprint]{neurips_2025}

\usepackage[utf8]{inputenc} 
\usepackage[T1]{fontenc}    
\usepackage{hyperref}       
\usepackage{url}            
\usepackage{booktabs}       
\usepackage{amsfonts}       
\usepackage{nicefrac}       
\usepackage{microtype}      
\usepackage{xcolor}         

\usepackage{graphicx}

\title{Human Semantic Representations of Social Interactions from Moving Shapes}

\author{%
  Yiling Yun \\
  Department of Psychology\\
  University of California, Los Angeles\\
  Los Angeles, CA 90095 \\
  \texttt{yiling.yun@g.ucla.edu} \\
 \AND
Hongjing Lu\\
  Department of Psychology, Department of Statistics\\
  University of California, Los Angeles\\
  Los Angeles, CA 90095 \\
  \texttt{hongjing@g.ucla.edu} \\
}

\begin{document}

\maketitle

\begin{abstract}
Humans are social creatures who readily recognize various social interactions from simple display of moving shapes. While previous research has often focused on visual features, we examine what semantic representations that humans employ to complement visual features. In Study 1, we directly asked human participants to label the animations based on their impression of moving shapes. We found that human responses were distributed. In Study 2, we measured the representational geometry of 27 social interactions through human similarity judgments and compared it with model predictions based on visual features, labels, and semantic embeddings from animation descriptions. We found that semantic models provided complementary information to visual features in explaining human judgments. Among the semantic models, verb-based embeddings extracted from descriptions account for human similarity judgments the best. These results suggest that social perception in simple displays reflects the semantic structure of social interactions, bridging visual and abstract representations.
\end{abstract}

\section{Introduction}
Humans are remarkably adept at recognizing social interactions. In the classic work by \citet{heider1944experimental}, observers who viewed a series of movements of simple geometric shapes reported intentions, emotions, and even personalities of these inanimate shapes. Subsequent studies using such simple stimuli of moving shapes showed that people from different cultures and age groups can reliably distinguish several evolutionarily relevant interactions (e.g., chase, fight) \citep{barrett2005accurate}. More recent work suggests this capacity extends to a broad set of socially meaningful actions including bothering and huddling \citep{roemmele2016recognizing}. Minimal animations of moving shapes elicit robust impressions of agency and interactions, despite lacking detailed body movements and being presented from an atypical bird's-eye perspective. For example, an animation that conveys ``talking'' contains no mouth movements. 

How do humans represent the wide range of social interactions from the minimum visual display of moving shapes? Prior work emphasizes visual cues that enable the perception of animacy and intentionality, including heading direction, relative speed, and acceleration \citep{dittrich1994visual, gao2011chasing}. Yet the perception of moving shapes can also be semantic, as observers spontaneously describe the shapes' behaviors with human-like terms, attributing actions, goals, and intentions. It is possible that semantic factors could contribute to the internal representation of social interactions. Previous research has analyzed human narrative descriptions of the animations to examine what events people describe and omit in order to evaluate how humans encode multi-agent movements \citep{wick2019perception}. This approach builds on the idea that observers condense complex motion into event-level descriptions, effectively filtering out extraneous details and revealing a more abstract representation of social interactions. 

Our study used animations of moving shapes presented in simple displays. Real-life social interaction videos would provide many cues including body postures, gaze directions, and facial expressions that can help interpret the interactions, but such richness does not explain how simple moving shapes can evoke the core of social representation. Additionally, we focus on the perception of interaction, which is the immediate impression after viewing a short animation, rather than on planning or instrumental reasoning across a sequence of actions. In the current study, we test the semantic representations that humans use for these human-human interactions displayed in the format of moving shapes. Study 1 measured observers' impressions of each animation. Study 2 measured the representational geometry of the animations with an odd-one-out task and compared the human representational geometry to multiple models: (i) low-level visual features, (ii) semantic labels, and (iii) text-derived embeddings from participants' descriptions. Since the animations depict actions with affective content, we focus on verbs and adjectives. Using this framework, we examine whether semantic features explain structure in human similarity judgments beyond what has been missed by visual features alone.

\section{Study 1: human impression of the moving shapes}
To measure human impression of the animations consisting of moving shapes, we conducted an empirical study where participants selected the best label that described each animation from a set of candidate labels. We also evaluated the behavior of a large language model (LLM) on this labeling task.

\subsection{Methods}
\subsubsection{Participants}
One hundred seventeen students in the Psychology department at UCLA participated in the online experiment for class credits. We excluded one participant who self-reported not being serious throughout the experiment, and two participants who self-reported not staying in the full-screen mode throughout the experiment. We analyzed data from the remaining 114 participants (Female: 91, Male: 21, Other: 1, Prefer not to say: 1; Mean age = 20.23). 

\subsubsection{Stimuli}
We used the Charade dataset \citep{roemmele2016recognizing} that contained animations of various social interactions depicted by two black triangles moving against a white background. One triangle was larger than the other one. The animations were created from a charade game where human annotators were asked to demonstrate action words (such as hug and fight) by manually moving the two triangles on a touchscreen. The trajectories and facing directions of the two triangles were recorded as the animations. The generated animations were further evaluated by a different group of participants. From this dataset, we selected 88 good-quality animations depicting 27 distinct social interactions with durations ranging from two to six seconds. For each interaction, we selected one animation that best described it. The selected 27 animations were labeled in the original dataset as follows: hug, huddle, kiss, approach, flirt, scratch, poke, creep, tickle, hit, talk, fight, escape, lead, herd, accompany, throw, ignore, leave, avoid, bother, push, capture, follow, pull, examine, and encircle. 

\subsubsection{Design}
The experiment was programmed in HTML, JavaScript, CSS, and PHP. On each trial, one animation was displayed at the center of the screen (300 pixels wide, preserving aspect ratio). After watching the animation, participants were instructed to ``[s]elect the label that best describes the animation'' from the 27 options, including hug, huddle, kiss, approach, flirt, scratch, poke, creep, tickle, hit, talk, fight, escape, lead, herd, accompany, throw, ignore, leave, avoid, bother, push, capture, follow, pull, examine, and encircle. Participants could replay the animation as many times as they wished; no feedback on their selection was provided. Each participant judged all 27 animations in random order. The display order of the label option set was randomized across participants but fixed within participants. The experiment lasted approximately 12 minutes.

\subsection{Results}
\begin{figure}[h]
    \begin{center}
        \includegraphics[width=0.95\linewidth]{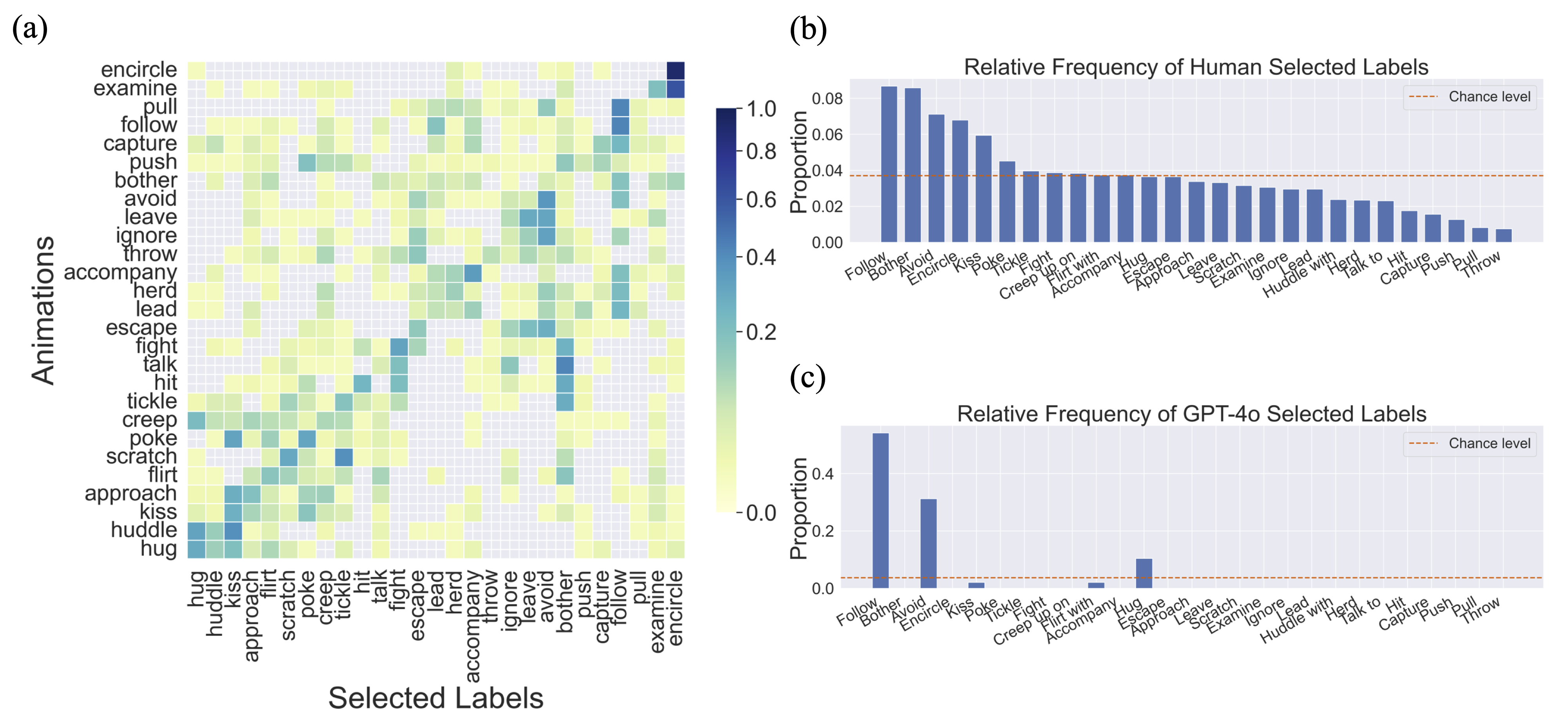}
    \end{center}
    \caption{(a) Response matrix of human judgments. Darker blue colors indicate a higher proportion of participants selecting the label as best describing the animation. Labels with no selection are shown in grey. A PowerNorm ($\gamma = 0.5$) transformation to the colormap scale was applied to enhance the visibility of small differences in low-proportion selections. (b) Relative frequency of label selections from human participants. (c) Relative frequency of label selections from GPT-4o. The response matrix from GPT-4o is shown in the appendix.}
    \label{fig:study1}
\end{figure}

For each animation, we measured the proportion of participants who selected each label in Figure~\ref{fig:study1}. The vertical axis referred to the animations in their original labels from the dataset, and the horizontal axis showed the response labels chosen by participants. The color of each cell represents the proportion of participants who chose that response label for the given animation. If all participants selected the response label for an animation that was consistent with its label in the original dataset, we would expect to see a diagonal line in the response-animation matrix. However, the figure revealed that most animations elicited multiple labels across different participants. Human selections were scattered than unanimous, indicating that participants had different impressions of the interaction from the same animation consisted of moving shapes. From the relative frequency of responses in Figure~\ref{fig:study1}, we observed that participants chose the labels ``follow'', ``bother'', ``avoid'', ``encircle'', and ``kiss'' well above the chance level of $\frac{1}{27}$.

We also compared human responses to label judgments made by GPT-4o. Because GPT-4o can only process at most 50 images at once, each of the 27 animations was sampled into frames at two rates: one every five frames and one every fifteen frames. The original animations were rendered at 50 frames per second, so these sampling rates preserved details for recognition. For each sampled sequence, the model was asked to select a label from the same set of 27 options and to provide a description of the animation. We evaluated the model's label selection based on the first output token. GPT-4o's label selections showed a narrower distribution as in Figure \ref{fig:study1}. The model most frequently selected ``follow'', ``avoid'', and ``hug'', which overlapped with human participants' common choices. These results indicate that GPT-4o exhibited a more biased labeling pattern compared to human participants. Overall, both humans and GPT-4o judged the animations in ways that diverged from the original dataset labels used to guide annotators in creating the animations.

\section{Study 2: representational geometry of social interactions through human similarity judgments}
The distributed impressions of the animations highlight the subtlety of social perception, since identical animations can be interpreted in different ways. We therefore ask how humans organize the variety of interactions in the mental space. In Study 2, we collected odd-one-out similarity judgments and converted them into a human dissimilarity matrix to represent the representational geometry. We then compared this human geometry to model distance matrices derived from visual features and language-based embeddings of participants' descriptions.

\subsection{Methods}
\subsubsection{Participants}
Seventy-seven students in the Psychology department at UCLA participated in the online experiment for class credits. We excluded one participant who did not complete all the trials, five participants who self-reported not being serious throughout the experiment, and one participant who self-reported not staying in the full-screen mode throughout the experiment. We analyzed data from the remaining 70 participants (Female: 57, Male: 13; Mean age = 20.49). 

\subsubsection{Design}
\begin{figure}[h]
    \begin{center}
        \includegraphics[width=0.8\linewidth]{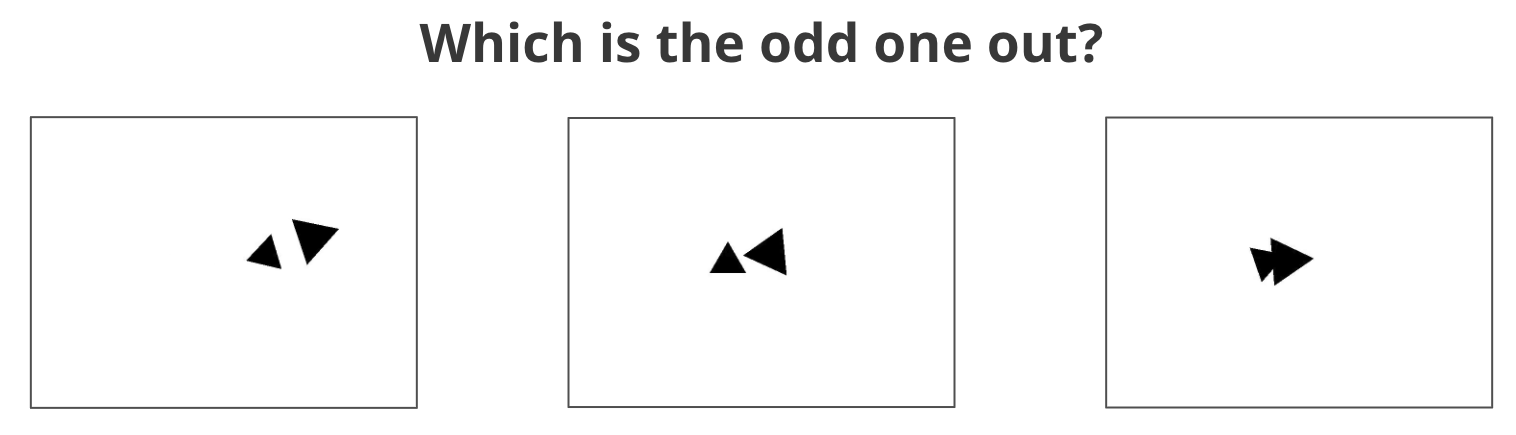}
    \end{center}
    \caption{Illustration of a static frame of each stimulus presented in one trial.} 
    \label{fig:study2_trial}
\end{figure}

We used the same 27 animations in Study 1. We used the odd-one-out task to assess the similarity between each pair of the 27 animations representing different social interactions. On each trial, we displayed three animations side by side in the center of the computer screen. Each animation was 300 pixels in width in its original ratio, and there was a gap of 20 pixels between each two of them (Figure \ref{fig:study2_trial}). After watching all three animations one after another, participants clicked on a button beneath the corresponding animation to indicate their odd-one-out judgment. When a participant selected one animation as the odd-one-out, their response implied that they considered the two unselected animations to be more similar to each other than to the selected animation. No feedback was provided, so participants were not guided to make judgments in a particular way. 

To test each pair of animations against all remaining 25 animations, we created the full combination of 2,925 unique trials and randomly assigned them to 65 different versions of the experiment. Each participant received one version. The order of the trials for each participant was randomized. The position of the three animations was also randomized for each trial. The experiment was programmed in HTML, JavaScript, CSS, and PHP. In total, each participant completed 45 trials over approximately 23 minutes. 

\subsection{Models}
\subsubsection{Low-level visual features}
Since the animations involve strong visual cues, we computed the following visual features for each animation: average speed, average acceleration magnitude, average velocity, average acceleration, duration, average relative distance, and average speed difference. We included the average speed and acceleration magnitude because entities' moving speed and the change in speed were salient. We included velocity and acceleration to take into account directions in addition to speed and acceleration magnitude. Average relative distance and average speed difference account for the difference between the two entities.

The specific calculations of speed and acceleration magnitude were as follows: To compute the average speed, the movement distance (displacement) of each triangle between two consecutive frames was calculated and then averaged over time across two triangles (unit: px/frame); average acceleration magnitude was calculated as the change in speed of each triangle in each frame and then averaged over time across two triangles. For each animation, there was one value for average speed and one value for average acceleration magnitude. The specific calculations of velocity and acceleration were as follows: to indicate the direction in velocity, we represented the average velocity of one triangle in a vector with two elements -- the average displacement in the horizontal and vertical direction over time for each triangle. We then concatenated the two vectors of the two triangles; for acceleration, we also represented it in a vector form with the average change in displacement in the horizontal and vertical direction over time for each triangle. We then concatenated the two vectors of the two triangles. Therefore, for each animation, there was one vector of four values for velocity and one vector of four values for acceleration. Duration was represented by one value. Average relative distance and average speed difference were calculated by taking the differences in location and speed between the two triangles, averaged over time; each was represented by a single value as well. Together, we concatenated all the features and generated a vector of length 13 for each animation. We then calculated the pairwise Euclidean distances to construct a distance matrix as an estimate of the dissimilarity judgments from these basic visual features.

\subsubsection{Semantic label}
We obtained word embeddings for the animation labels from the Charade dataset using the fastText model \citep{bojanowski2016enriching}. Each social interaction's label was represented by a vector of length 300. We used pairwise cosine distance to estimate the dissimilarity matrix.

\subsubsection{Semantics from descriptions}
In a separate study, we collected human descriptions of the animations (N = 39). Each participant described all 27 animations. We then obtained embeddings of their descriptions using the SentenceTransformer model \citep[all-mpnet-base-v2, ][]{reimers-2019-sentence-bert} in vectors of length 768. Specifically, we tokenized and parsed each description with spaCy \citep[i.e., the small English model without the Named Entity Recognizer component for simplicity, ][]{Honnibal_spaCy_Industrial-strength_Natural_2020}. Text was lower-cased and stripped of extra whitespace. We used spaCy's POS tags and dependency relations \citep{Honnibal_spaCy_Industrial-strength_Natural_2020} to locate sentences, verbs, and adjectives. Each lemma was embedded independently, and the description's embedding vector was the pooled mean of the lemma embeddings. L2 normalization was applied before and after pooling. For each participant's descriptions, we calculated pairwise cosine distances across the 27 animation embeddings, and then averaged these matrices across participants to derive the final distance matrix.

Using the above pipeline, we tested four models: sentence, verb-only, verb-xor-adjective, and verb-and-adjective. In the \textbf{sentence} model, we input all of the human descriptions to the SentenceTransformer. In the \textbf{verb-only} model, we assumed that verbs were the most crucial part of human descriptions, extracted lemmas of tokens tagged VERB in the descriptions, and obtained the embeddings based on the verbs. If a description contained no verb (e.g., ``extremely aggressive behavior''), we embed the full sentence description as a fallback. The \textbf{verb-xor-adj (V$\oplus$A)} model was inspired by some descriptions that did not include any verb but with adjectives. In V$\oplus$A, we still prioritized the verbs in the descriptions. Only if the description did not contain a verb, we extracted the lemmas of tokens tagged ADJ as a fallback. If the description contained neither a verb nor an adjective, we obtain its sentence embedding. In the \textbf{verb-and-adj (V+A)} model, we obtained embeddings from all verbs and adjectives in the descriptions. When getting adjectives from the verb-or-adj and verb-and-adj, we ignored the words ``big'' and ``small'' because they were used to refer to either triangle instead of describing the interactions between them.

\subsection{Results}
\begin{figure}[h]
    \begin{center}
        \includegraphics[width=0.9\linewidth]{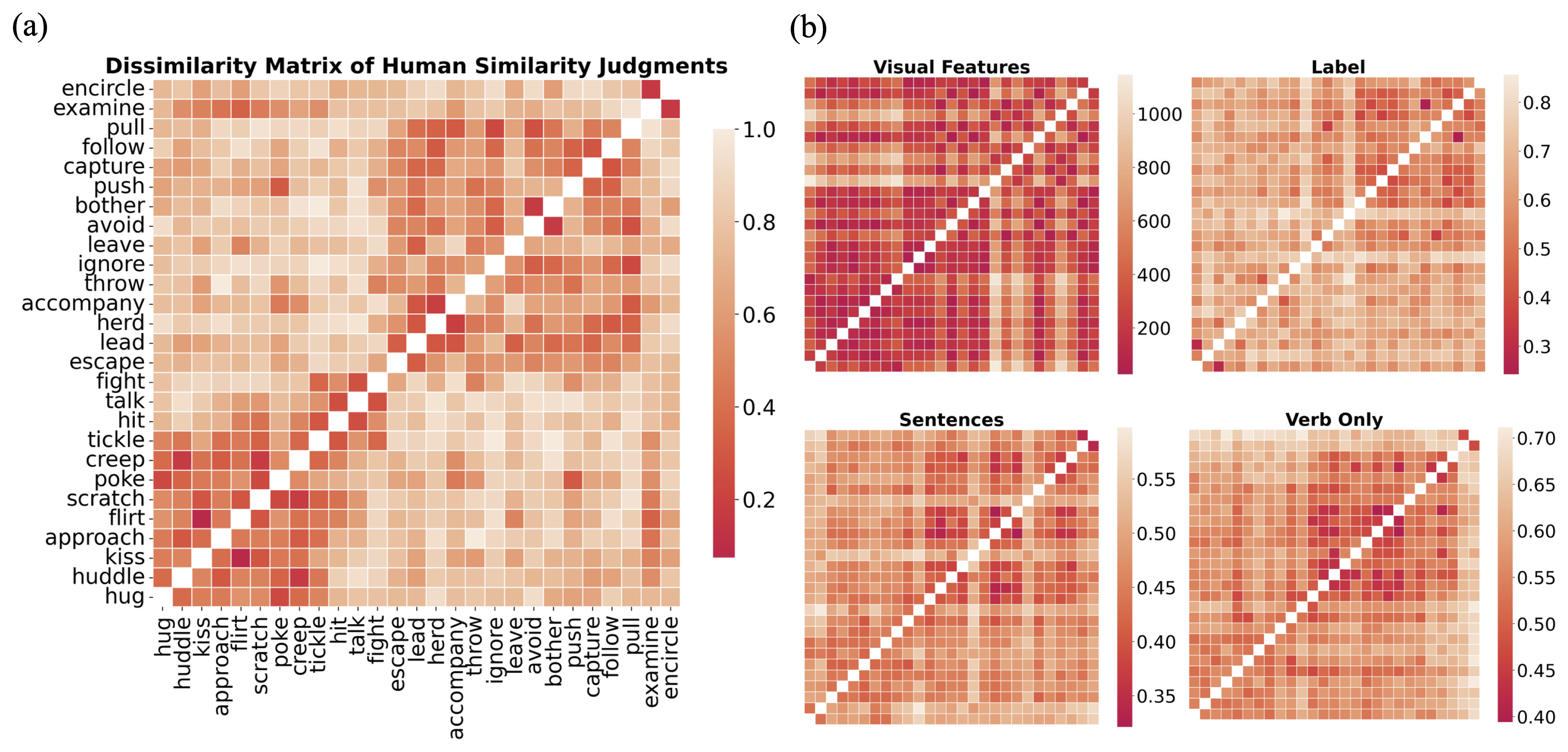}
    \end{center}
    \caption{(a) Dissimilarity matrix of human judgments. The darker red indicates a higher similarity (low dissimilarity) between the two corresponding animations. (b) Predicted dissimilarity matrices from selected models. The V$\oplus$A and V+A model results were shown in the appendix. The horizontal and vertical labels are the same as in (a).} 
    \label{fig:study2_dist_mat}
\end{figure}

From the behavioral experiment, we calculated the dissimilarity scores based on the odd-one-out task. We first calculated the similarity scores that were determined by the proportion of trials in which neither animation was chosen, considering all trials that included the given pair. We then subtracted the similarity scores from 1 to obtain the dissimilarity scores. Figure~\ref{fig:study2_dist_mat} illustrates the human dissimilarity scores, revealing distinct similarity structures across various types of social interactions. Two clusters emerged from the representational geometry: one in the bottom left involving actions where two shapes come close (e.g.,  ``hug'' and ``approach''), and the other in the top right involving at least one shape moving away from the other (e.g., ``throw'' and ``ignore''). The reliability noise ceiling calculated through the split-half of human data was 0.811 ($p< .001$; $CI = [0.785, 0.837]$), which represents the highest correlation possible with the given human similarity judgments for computational models. 

\begin{figure}[h]
    \begin{center}
    \includegraphics[width=0.8\linewidth]{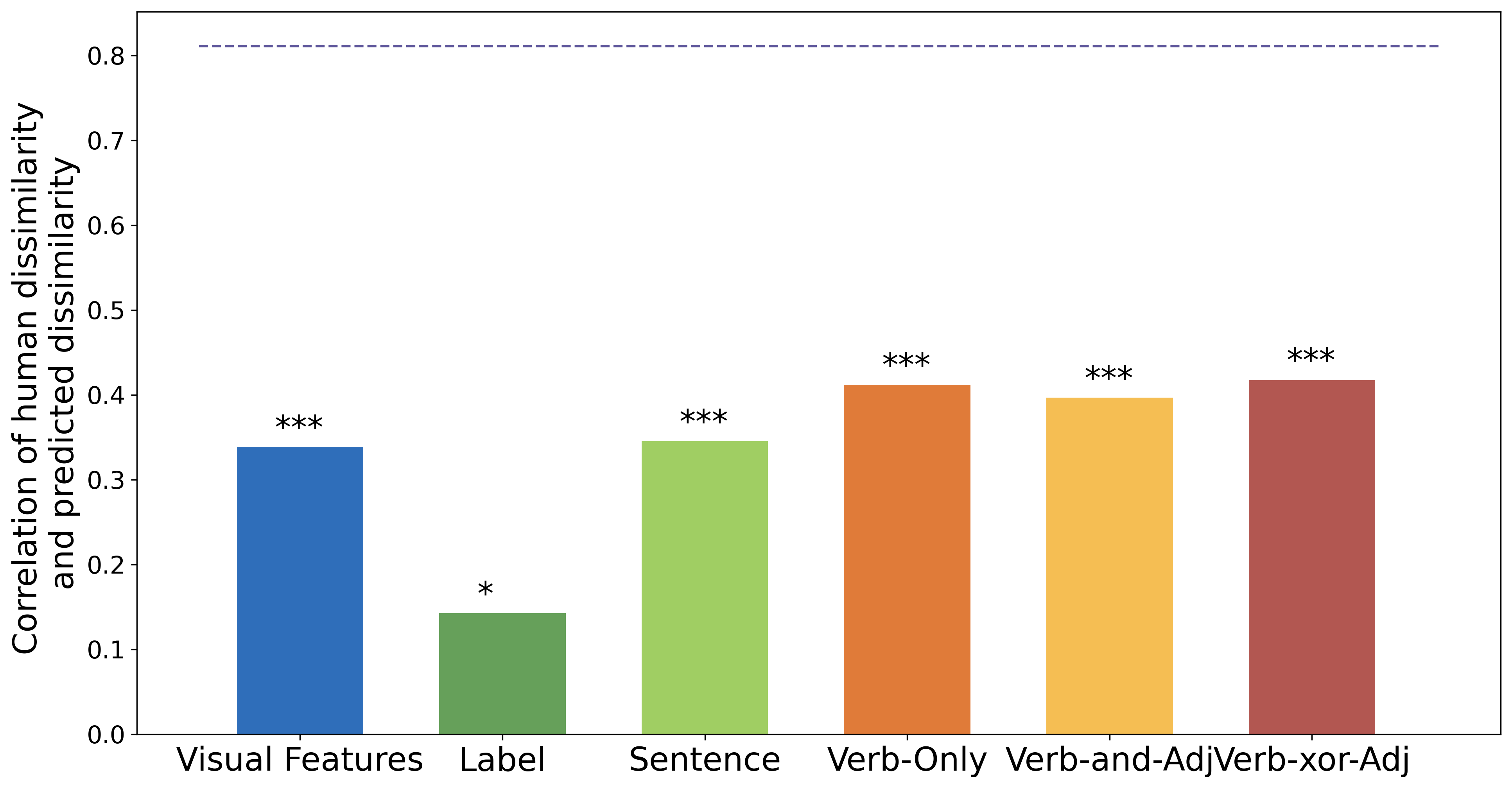}
    \end{center}
    \caption{Correlation between human dissimilarity judgments and model-predicted dissimilarity of animations.} 
    \label{fig:study2_corr}
\end{figure}

We then compared the human similarity judgments with the modeling results (Figure~\ref{fig:study2_dist_mat}) by computing the Pearson correlation between human dissimilarity scores and model-predicted distance scores. Figure~\ref{fig:study2_corr} shows the result between model predictions and the human similarity judgments: The visual features generated moderate correlations ($r = .338$). Semantic labels from the original dataset showed the least correlation ($r = .143$). Among the language embeddings based on human descriptions, the verb-xor-adj model and the verb-only model correlated with human similarity judgments the highest ($r = .417$ and $r = .412$ respectively), followed by the verb-and-adj model and the sentence model ($r = .396$ and $r = .345$ respectively). All correlations were significant (semantic label showed $p = .011$, other models showed $p$s $< .001$, corrected through Mantel's test, \citep{mantel1967detection}).

Using a semi-partial correlation test to control the effect of all language models on predicting human similarity judgments, the visual features provided unique contribution to account for human judgments, as shown by a significant semi-partial correlation ($sr = .352$, $p < .001$). Among the language models from human descriptions, the V$\oplus$A model was the minimal representation, capturing human similarity ratings with the fewest linguistic features. Controlling for the V$\oplus$A model on predicting human similarity judgments, labels from the dataset alone did not have a significant contribution anymore ($sr = .039$, $p = .468$). When controlling for the verb-only model, V+A explained no additional variance in human similarity judgments ($sr = .009$, $p = .873$). Conversely, when controlling for V+A, the verb-only model explained no additional variance either ($sr = .041$, $p = .441$). The comparison between the verb-only model and V$\oplus$A showed the same pattern ($sr$s $< .01$, $p$s $>.80$). The verb-only model, V+A model, and V$\oplus$A model explained overlapping variance in human similarity judgments, which indicates that the explained variance was primarily driven by verb embeddings, and adjectives contribute negligible explanatory ability. Controlling for sentences on human similarity judgments, the verb-only model showed a unique contribution ($sr = 0.233$, $p < .001$). These results show that language models and visual features capture for different aspects of human similarity judgments. Language embeddings derived from human descriptions correlated with human similarity judgments better than fastText embeddings of labels from the original dataset. Lastly, verbs appeared critical for explaining human similarity ratings and provided a more parsimonious account than full-sentence descriptions.

\section{General discussion}
Studying human impressions of social interactions with highly simplified stimuli allows us to focus on the core aspects of social interaction representations. Study 1 showed that the interpretations of the animations were distributed across human participants. Study 2 examined the human representational geometry of a variety of interactions and tested it against different models. We found that language models and visual features explain complementary aspects of human representations. Among language models, embeddings derived from participants' descriptions correlated more strongly with human judgments than embeddings of the category labels alone. Within the description-based models, verb-focused embeddings provided the strongest fits to human similarity structure. While the current study focused on 27 interactions, future studies could include more instances of each interaction.

The findings that some actions are more frequently selected by both humans and LLM, and that description-based embeddings outperform original label embeddings, shed light on future research directions concerning the hierarchy of interaction concepts. Some action verbs might capture basic interactions, while others reflect compositions of these interactions. Related work in event segmentation indicates that inferred intentions can modulate kinematic features at event boundaries \citep{zacks2004using}, which hints at a layered representation in which semantics guides which features are treated as relevant. Finally, while the present data suggest that adjectives contribute limited unique variance beyond verbs in this paradigm, action manners that are often expressed with adverbs remain a promising target for future work to capture the fine-grained nuances of social interactions. Overall, this work unravels how people represent social interactions beyond visual features and provides a foundation for further explanatory models to better understand social cues, while also recognizing that such research may be vulnerable to misuse in manipulative or surveillance applications.

\begin{ack}
This work was supported by the NSF grant BCS-2142269.
\end{ack}

\bibliographystyle{apalike}
{
\small
\bibliography{bibliography}



%
}


\appendix

\section{Technical Appendices and Supplementary Material}
\subsection{Study 1 procedure}
Participants accessed the experiment from their personal laptops. After reading the instructions, they completed one practice trial and a comprehension check. They then provided informed consent to participate in the study. At the end of the study, participants reported whether they had remained attentive (e.g., serious and in full-screen mode throughout the experiment), and answered questions about confusion, comments, age, and gender. 

\subsection{Study 1 GPT-4o}
The prompt for GPT 4o was ``Select the label that best describes the animation.

Options: huddle with, escape, lead, herd, accompany, ignore, bother, capture, follow, pull, encircle, push, tickle, scratch, kiss, hug, avoid, flirt with, throw, approach, poke, examine, creep up on, talk to, fight, leave, hit, argue with, chase, mimic, and play with. 

Respond in the following format exactly in two lines:

<word label from the list> - no punctuation, no formatting, no explanation.

Description: <your description here>
''

We ensured that the input image filenames followed the format of ``frame\_\#\#\#\# (four-digit numbering)'' without including any action labels in the filename. The inputs were downsampled JPEG images at one-quarter of the original resolution. This choice was motivated by the simplicity of the stimuli in that the images contained mostly a white background with two black triangles and thus required minimal visual detail.

When sampling every five frames, six animations could not be processed and returned errors due to exceeding the input length limit. In the model outputs, the first token occasionally appeared as ``fl'', which we translated to flirt with, as it was the only option label beginning with those characters.

\begin{figure}[h]
    \begin{center}
    \includegraphics[width=0.4\linewidth]{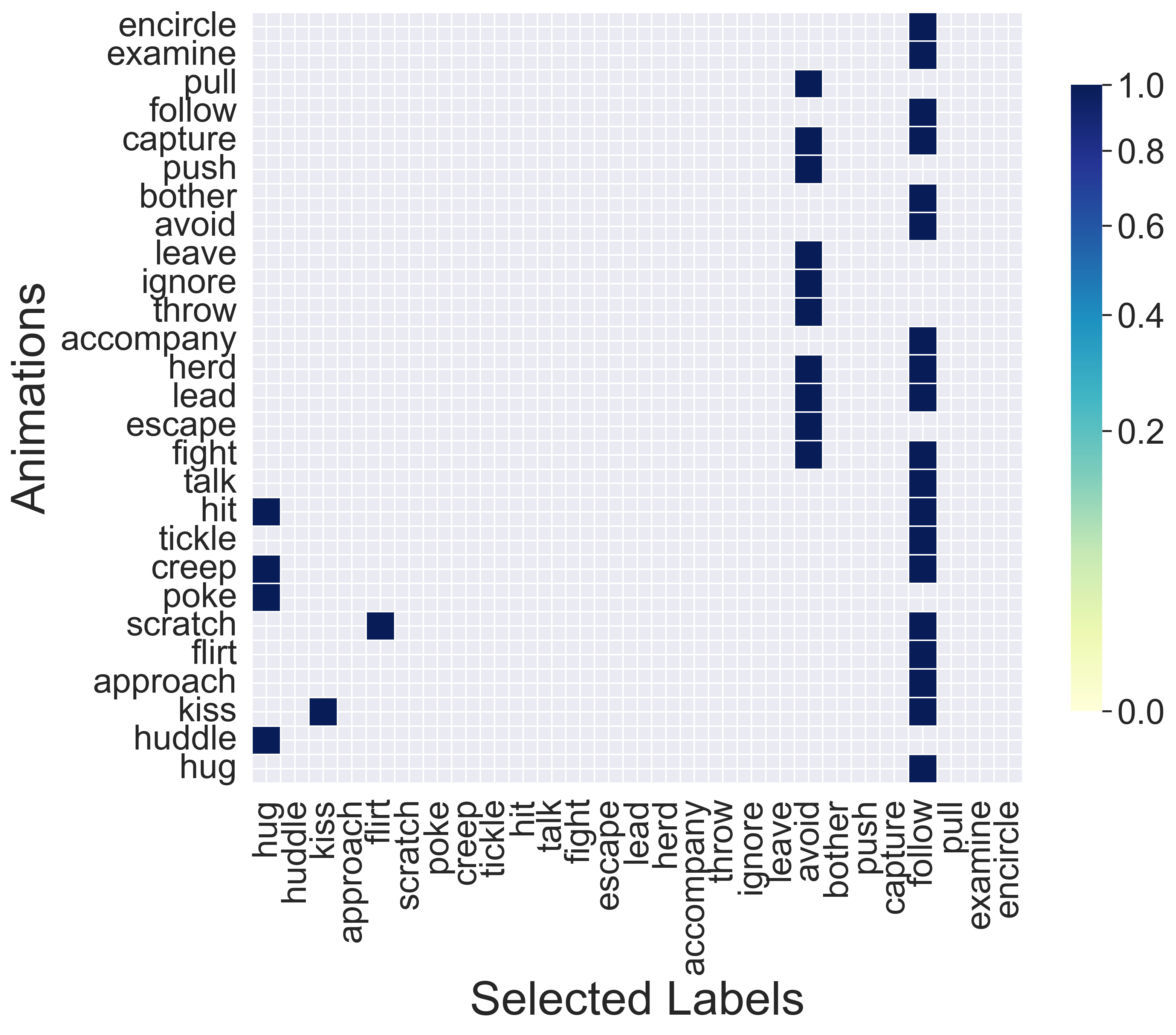}
    \end{center}
    \caption{Response matrix of GPT-4o.} 
    \label{fig:study1_gpt_conf_mat}
\end{figure}

\subsection{Study 2 odd-one-out procedure}
Participants accessed the experiment from their personal laptops. They first read the instructions about the task and were shown an example animation of pushing. They then familiarized themselves with the task through one example trial of the odd-one-out task. After an instruction quiz question that tested their understanding of the task, they gave consent to start the experiment. There was no time limit for their decisions. There was a progress bar at the top of the screen. Participants could watch the animations in any order and for as many times as they wanted. They could only proceed to the next trial after they had watched all three animations and selected one odd-one-out animation. After completing all the trials, we administered some survey questions to ask if they were serious and in full-screen mode throughout the experiment, had any comments about the study, and had encountered any technical issues. 

\subsection{Study 2 embeddings based on human descriptions}
The study to collect human descriptions involved two parts. In the first part, participants rated to what extent the two triangles were socially interacting on a Likert scale from 1 to 5. In the second part, on each trial, participants first selected the label that best described the animation and then explained why the selected label best described the animation. The explanatory component of the second part of the study provided the descriptions used in the embedding analyses.

\begin{figure}[h]
    \begin{center}
    \includegraphics[width=0.4\linewidth]{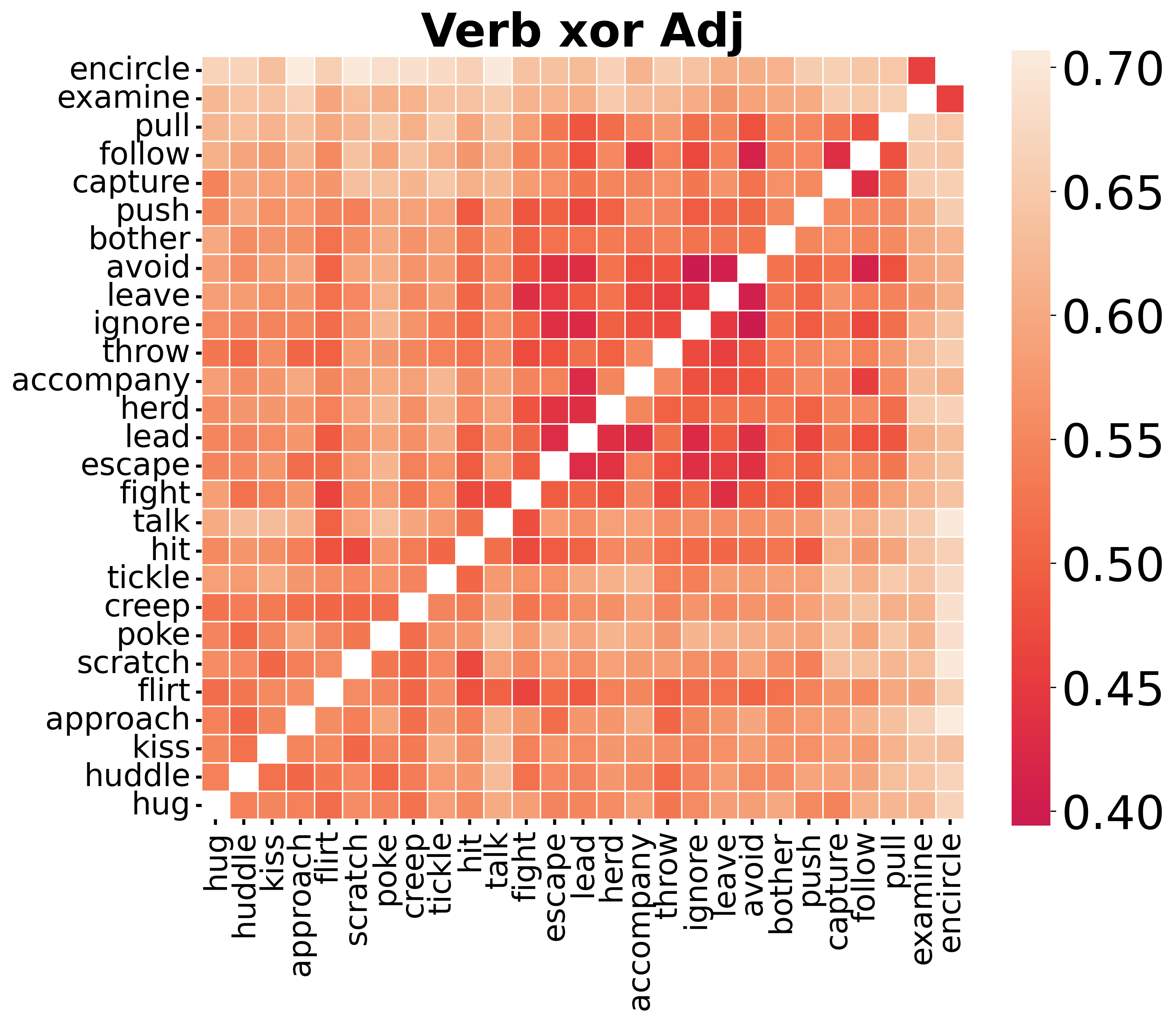}
    \includegraphics[width=0.4\linewidth]{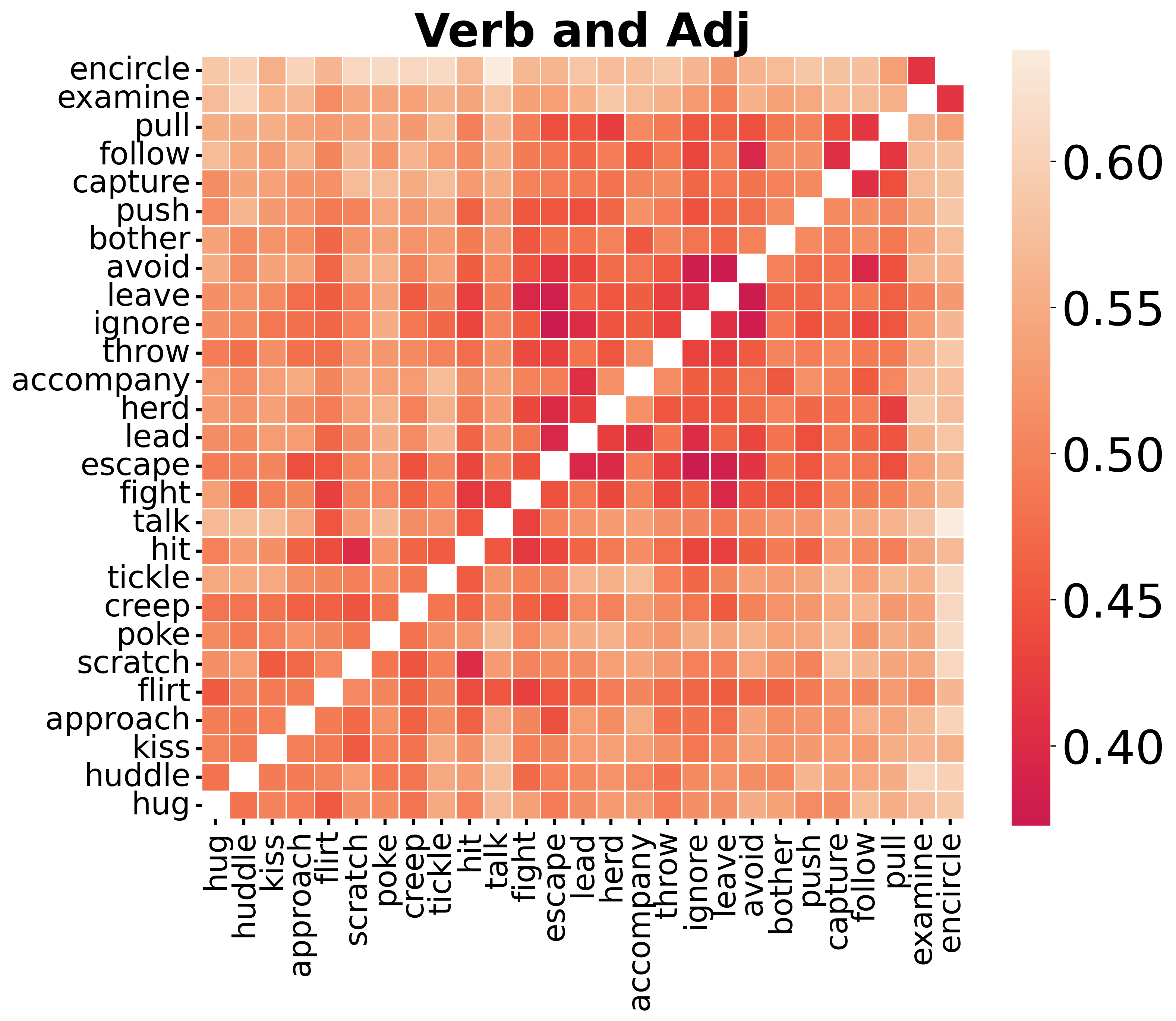}
    \end{center}
    \caption{Predicted dissimilarity matrix of human judgments from the Verb-xor-Adj and Verb-and-Adj models.} 
    \label{fig:study2_ooo}
\end{figure}


\subsection{Ethics}
The experiments posed no potential risks on participants and were performed in accordance with guidelines and regulations approved by the institutional review board (IRB).

\end{document}